\documentclass[lettersize,journal]{IEEEtran}
\usepackage{amsmath,amsfonts}
\usepackage{algorithmic}
\usepackage{array}
\usepackage[caption=false,font=normalsize,labelfont=sf,textfont=sf]{subfig}
\usepackage{textcomp}
\usepackage{stfloats}
\usepackage{url}
\usepackage{verbatim}
\usepackage{graphicx}
\usepackage{cite}
\usepackage{amssymb}
\usepackage{bbold}
\usepackage[ruled,vlined]{algorithm2e}
\usepackage{enumitem}

\usepackage{url}

\usepackage{hyperref} 

\begin{document}
\title{Example Forgetting: A Novel Approach to Explain and Interpret Deep Neural Networks in \\Seismic Interpretation}

\author{{Ryan Benkert, \IEEEmembership{Student Member, IEEE}, Oluwaseun Joseph Aribido, \IEEEmembership{Student Member, IEEE}, \\and Ghassan AlRegib, \IEEEmembership{Senior Member, IEEE}}
\thanks{Manuscript received February 08, 2022.}}

\markboth{Accepted for Publication in IEEE Transactions on Geoscience and Remote Sensing May~2022}%
{Shell \MakeLowercase{\textit{et al.}}: A Sample Article Using IEEEtran.cls for IEEE Journals}


\onecolumn 

\begin{description}[labelindent=-1cm,leftmargin=1cm,style=multiline]

\item[\textbf{Citation}]{R. Benkert, O.J. Aribido, and G. AlRegib, "Example Forgetting: A Novel Approach to Explain and Interpret Deep Neural Networks in Seismic Interpretation," in IEEE Transactions on Geoscience and Remote Sensing (TGRS), May. 12 2022} \\


\item[\textbf{Review}]
{
Date of submission: October 2021\\
Date of acceptance: May 2022
} \\


\item[\textbf{Bib}] {@ARTICLE\{benkert2022\_TGRS,\\ 
author=\{R. Benkert, O.J. Aribido, and G. AlRegib\},\\ 
journal=\{IEEE Geoscience and Remote Sensing\},\\ 
title=\{Example Forgetting: A Novel Approach to Explain and Interpret Deep Neural Networks in Seismic Interpretation\}, \\ 
year=\{2022\}\\ 
} \\


\item[\textbf{Copyright}]{\textcopyright 2022 IEEE. Personal use of this material is permitted. Permission from IEEE must be obtained for all other uses, in any current or future media, including reprinting/republishing this material for advertising or promotional purposes,
creating new collective works, for resale or redistribution to servers or lists, or reuse of any copyrighted component
of this work in other works. }
\\
\item[\textbf{Contact}]{\href{mailto:rbenkert3@gatech.edu}{rbenkert3@gatech.edu}  OR \href{mailto:alregib@gatech.edu}{alregib@gatech.edu}\\ \url{http://ghassanalregib.info/} \\ }
\end{description}

\thispagestyle{empty}
\newpage
\clearpage
\setcounter{page}{1}

\twocolumn

\maketitle

\begin{abstract}
In recent years, deep neural networks have significantly impacted the seismic interpretation process. Due to the simple implementation and low interpretation costs, deep neural networks are an attractive component for the common interpretation pipeline. However, neural networks are frequently met with distrust due to their property of producing semantically incorrect outputs when exposed to sections the model was not trained on. We address this issue by explaining model behaviour and improving generalization properties through example forgetting: First, we introduce a method that effectively relates semantically malfunctioned predictions to their respectful positions within the neural network representation manifold. More concrete, our method tracks how models "forget" seismic reflections during training and establishes a connection to the decision boundary proximity of the target class. Second, we use our analysis technique to identify frequently forgotten regions within the training volume and augment the training set with state-of-the-art style transfer techniques from computer vision. We show that our method improves the segmentation performance on underrepresented classes while significantly reducing the forgotten regions in the F3 volume in the Netherlands.
\end{abstract}


\section{Introduction}
\begin{figure}[!h]
\begin{center}
\includegraphics[scale=0.5]{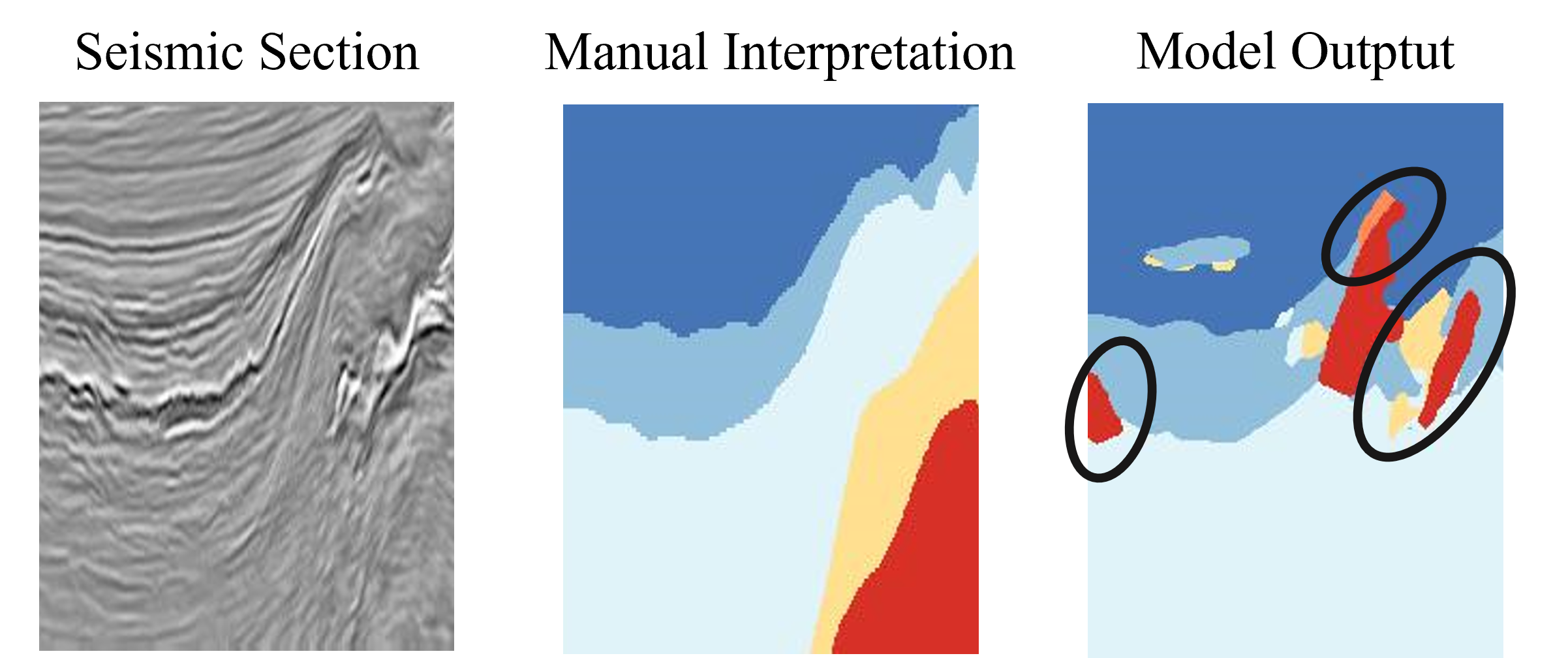}
\caption{An example of mispredictions of a deep neural network. Left: Seismic input section. Center: Manual seismic interpretation. Right: Automated interpretation of a trained deep neural network. The circled sections represent severely mispredicted regions.}
\label{fig:mispredicitons-example}
\end{center}
\end{figure}

\IEEEPARstart{I}{n} the field of geophysics, interpreting processed seismic images is a challenging task. For decades, the process required expert oversight and a costly interpretation process. The introduction of deep learning to the field of geophysics significantly sped up this task and enabled accurate interpretation with limited human interference \cite{8312469}. Instead of experts annotating volumes for weeks, the interpreter trains a deep model on a annotated training volume and subsequently infers geological information from similar test volumes in a matter of hours. The reason for this success is tied to the nature of the interpretation task. Traditionally, the interpreter extracts quantitative measures (attributes) of interesting characteristics and infers geological information based on the extracted attributes and the seismic section \cite{chopra2005seismic}. The choice of these attributes depends on the interpretation objective. For instance, several attributes are based on geometric properties \cite{taner1994seismic}, \cite{barnes1992calculation}, \cite{chen1997seismic} while others are derived from the human visual system \cite{shafiq2015detection, shafiq2017salt, shafiq2018role}. At its core, deep models function in a very similar fashion. The model extracts complex features from the seismic section and classifies the features based on the trained loss objective. By construction, convolutional neural networks \cite{lecun1995convolutional} posses the capability to model complex spatial features that human interpreters may overlook or that hand-engineered attributes may not capture. This specific characteristic is one of the biggest advantages of deep neural networks, but can also be a huge pitfall. On one hand, deep models automate the attribute extraction process and model complex seismic features easily omitted by interpreters or geometric attributes. This relieves the interpreter from selecting the appropriate attribute and significantly decreases the overall interpretation time. On the flip side, deep models lack interpretability. Even though the model automates feature extraction, it is unclear how these features are related to the semantic interpretation of the subsurface. In other words, the interpreter is unable to explain the behavior of the network because the features are not necessarily based on geophysical information. In many cases, this leads to unexplainable predictions that undermine the confidence in deep models during inference. For instance, a trained machine learning model may predict deep subsurface structures as near surface facies (Figure~\ref{fig:mispredicitons-example}). The former interpretation is significantly less likely in the traditional workflow where attributes are based on relevant geophysical characteristics and interpretations are performed manually by humans.\\

In this paper, our objective is to make interpretations more explainable and predict the behavior of deep models when utilized on different input volumes. In the field of computer vision, several approaches to model uncertainty or interpretability involve Bayesian inference \cite{kendall2017uncertainties, gal2016dropout} or gradient-based approaches \cite{selvaraju2017grad, lee2020gradients, prabhushankar2020contrastive, kwon2020backpropagated, kwon2020novelty}. Even though these approaches are useful visualization techniques, they fail to provide information about the relationship of samples with respect to the decision boundary. In seismic interpretation, this information is especially important as it can provide useful insights about the geophysical relationship of the extracted features. Based on these observations, we propose using example forgetting to explain seismic deep models. At its core, our method tracks the frequency in which amplitude reflections are forgotten during training and highlights difficult regions in heat maps. From an optimization standpoint, our algorithm contrasts difficult samples with frequent decision boundary shifts from less difficult samples consistently mapped within class manifolds. This provides the interpreter with a powerful tool to visualize the generalization capabilities of the model and to verify model generalization with respect to the interrpetation of the subsurface.\\

In summary, our contributions in this paper are as follows: First, we present a framework to \emph{explain deep model behavior} by evaluating the learning dynamics during training. Second, we \emph{analyze deep models} by visualizing challenging regions and interpret them with respect to model predictions and the geophysical properties of the subsurface. The framework allows us categorize prone regions and evaluate their contribution to the model performance. Third, we \emph{introduce a segmentation framework} that explicitly targets forgotten samples and significantly reduces difficult pixels. Our empirical findings show that our method impacts the representation space mapping and increases the distance of pixels to their respective class decision boundary. Moreover, our framework improves segmentation performance of underrepresented classes.
\section{Related Work}
In seismic interpretation, deep learning models first surfaced in the form of fully supervised settings \cite{di2018patch, wu2018convolutional}. However, due to the high cost, fully annotated datasets are scarce in seismic interpretation. For fully supervised models, this frequently causes overfitting and poor prediction capabilities. As a result, several works explored semi-supervised and weakly supervised approaches \cite{alaudah2016weakly, alaudah2018structure, alaudah2019facies, alaudah2018learning, alfarraj2019semisupervised, babakhin2019semi} as these methods are less dependent on costly data annotations. Apart from methodological shifts, deep models have been further diversified on different seismic applications. A few example applications include detection of faults \cite{araya2017automated, di2019improving, di2019semi, shafiq2018novel,wu2019faultseg3d,xiong2018seismic}, delineation of salt bodies \cite{di2018deep, di2018multi, shafiq2015detection}, classification of facies \cite{liu20193d, dramsch2018deep, qian2017seismic, alaudah2019facies, alaudah2019machine}, prediction rock lithology from well logs \cite{alfarraj2019semisupervised, das2018convolutional, das2019effect, ahmadjournal} and seismic horizon interpretation \cite{tschannen2020extracting, wu2019semiautomated}.\\

Although deep learning models are effective, they are hard to explain and mispredictions may follow a random pattern that is semantically incorrect. In computer vision, this is a well known issue of deep models and works on model uncertainty or explainability are ubiquitous. For instance, one branch focuses on utilizing gradient activations to infer information about the expected change a model witnesses when updating weight parameters \cite{lee2020gradients, prabhushankar2020contrastive, kwon2020backpropagated, kwon2020novelty}. A more traditional approach is visualizing model uncertainty through Bayesian inference \cite{kendall2017uncertainties, gal2016dropout}. Typically, this involves estimating the posterior probability of model parameters with respect to given data samples and their respective labels. Subsequently, model uncertainty is visualized by sampling from the parameter distribution and computing the entropy of the resulting prediction distribution. In seismic interpretation, most approaches concerning model uncertainty fall into this area of research \cite{zhao2020enrich, choi2020uncertainty, mukhopadhyay2019bayesian}.\\

In contrast to existing methods, we explain model behaviour and prediction uncertainty by investigating the learning dynamics in neural networks. In literature, research in this area can be broadly classified in two categories: The first category explores the learning continuity when deep models are trained on new tasks. In research, this behaviour is often referred to as catastrophic forgetting \cite{ritter2018online, kirkpatrick2017overcoming}. In seismic interpretation, we frequently encounter this phenomenon in transfer learning scenarios where models are pretrained on one dataset and fine-tuned on another. The second category addresses the learning behaviour within a single task and analyzes sample forgetting within the training distribution \cite{toneva2018empirical}. In this paper, we generalize this concept to a seismic segmentation problem and visualize frequently forgotten regions in heat maps. Further, we exploit frequently forgotten regions by transferring their class characteristics to different sections within the seismic volume.  To achieve this, we utilize state-of-the-art style transfer algorithms from computer vision.\\
In this context, several style transfer approaches are based on conditional generative adversarial networks \cite{mirza2014conditional}. Starting with \cite{isola2017image}, conditional generative adversarial networks (cGANs) have been widely deployed in many image-translation applications due to their high quality image generation characteristics. Examples of such applications are high resolution image synthesis \cite{wang2018high}, multi-modal image synthesis \cite{zhu2017toward, huang2018multimodal} and semantic image synthesis \cite{zhu2020sean, bau2020semantic, gu2019mask, lee2020maskgan, park2019semantic}. In seismic, style transfer does not have much research traction. However, few papers address the generation of synthetic subsurface models by applying style transfer techniques \cite{feng2020physically, ovcharenko2019style}.

Finally, we note that this work is a continuation of \cite{benkert2021explaining}. In addition to \cite{benkert2021explaining}, we present significantly improved segmentation results as well as thorough analysis aspects of our method.

\begin{figure*}[!ht]
\begin{center}
\includegraphics[scale=0.6]{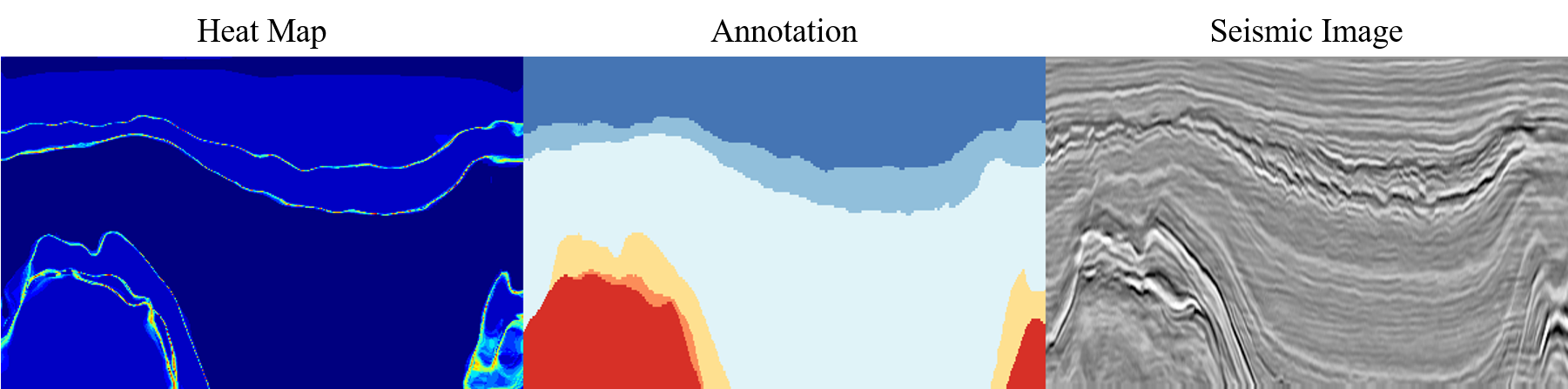}
\caption{An example of our proposed method. Left: Heat map of the section. Pixels close to the decision boundary are highlighted in different shades of red whereas pixels deep within the class manifold are dark blue. Center: Annotation of the respective section. Right: Original seismic section. }
\label{fig:example-forgetting-example}
\end{center}
\end{figure*}
\section{Explainability in Neural Networks with Example Forgetting}
\label{sec:forgetting-events-test}
At the core of our technique stands the concept of example forgetting. Intuitively, samples that are more difficult to learn exhibit different properties than samples that are easy to distinguish and classify. In this section, we formalize this concept in the form of "forgetting".


\subsection{Forgetting Events}
Deep neural networks cannot learn continually but forget samples during the optimization process. More generally, optimizing weight parameters causes a shift in the representation manifold that can result in misprediction (or "forgetting") of previously correct samples. In neural networks, a shift occurs when a sample has been "learnt" (classified correctly) at some point $t$ and subsequently "forgotten" (misclassified) at a time $t' > t$. \\
Formally, we define $(x_i, y_i) \in I^{M\times N}$ as a (pixel, annotation) tuple in image $I$, where $x_i$ and $y_i$ correspond to the pixel and annotation respectively. In image segmentation, our goal is to calculate a prediction $\tilde{y}_i$ such that $\tilde{y}_i = y_i$. Based on this definition, the accuracy of a pixel in training epoch $t$ is defined as
\begin{equation}
acc^{t}_{i} = \mathbb{1}_{\tilde{y}^t_i = y_i}.
\end{equation}
Here, $\mathbb{1}_{\tilde{y}^t_i = y_i}$ refers to a binary variable indicating the correctness of the classified pixel in image $I$. With this definition we say a pixel is forgotten at epoch $t+1$ if the accuracy at $t+1$ is strictly smaller than the accuracy at epoch $t$:
\begin{equation}
f_{i}^{t} = int( acc^{t+1}_{i} < acc^{t}_{i} ) \in {1, 0}
\end{equation}
Similar to \cite{toneva2018empirical}, we define the binary event $f_i^t$ as a \emph{forgetting event} at time $t$.\\

In contrast to other deep learning applications, the nature of the segmentation tasks enables visualization of forgetting events (Figure~\ref{fig:example-forgetting-example}). Following our previous definition, we visualize forgetting events in a heat map by counting the number of forgetting events $f_i^t$ that occur per pixel during the time frame $T$. Mathematically, heat map $L \in \mathbf{N_{0+}}^{M\times N}$ and every element in $L$ can be written as

\begin{equation}
L_{i} = \sum_{t = 0}^{T}{f_{i}^{t}}.
\end{equation} 

Since frequently forgotten samples were shifted over the decision boundary frequently during training, we interpret forgetting events as an approximate metric for decision boundary proximity. This view is complementary to \cite{toneva2018empirical} where frequently forgotten samples are considered support vectors within the representation space. Qualitatively, we note that frequently forgotten regions typically contain overlapping class features or a significant amount of annotation ambiguity. For this reason, we mostly find forgettable regions in underrepresented classes (e.g. salt domes) or facies boundaries where annotations are the most ambiguous (Figure~\ref{fig:example-forgetting-example}).

\subsection{Heat Map Computation}
In our implementation, we calculate heat maps for different distribution sets. Following the previous definitions, we would have to track forgetting events for each model update. Practically, this would result in interpreting every volume set after each minibatch and updating the heat maps multiple times every epoch. Since this approach is computationally expensive, we update the heat maps of the current minibatch only. For the training set we monitor forgetting events of each minibatch and update with the corresponding batch gradient. For validation and test sets we track forgetting events after each epoch. Algorithm~\ref{alg:learningmapcomputation} outlines the tracking procedure. During training, we count the number of forgetting events for each pixel $(i, j)$ in set $D$ and store the result in a heat map for each image within the minibatch. If $D$ is the training set, we further update with the minibatch $B$. In all other cases (e.g. test), we do not perform model updates since this would alter the regular training procedure. Instead, we train the model for another epoch on the training set synchronously.\\

\begin{algorithm} [h]
\SetAlgoLined
\KwResult{Heat Maps of Set $D$.}
 Let $(i, j) \in M\times N \ \text{and}\ k \in D$\; 
 initialize $previousacc_{i, j}^k = 0$\;
 initialize Heat Maps $T_{i, j}[k] = 0$\;
 \While{Training not finished}{
  \For{$\text{batch } B \in D$}{
  \If{$D$ is training set}{
    update segmentation model on $B$\;
  }
    \For{$\text{pixel } (b, i, j) \in |B| \times M\times N$}{
        \eIf{$acc_{b, i, j}^k < previousacc_{b, i, j}^k$}{
        $T_{b, i, j}[k] += 1$\;
        }{
        continue\;
        }
    }
  }
 }
 \caption{Heat Map Computation}
 \label{alg:learningmapcomputation}
\end{algorithm}

\section{Support Vector Augmentation}
\begin{figure}
\begin{center}
\includegraphics[scale=0.6]{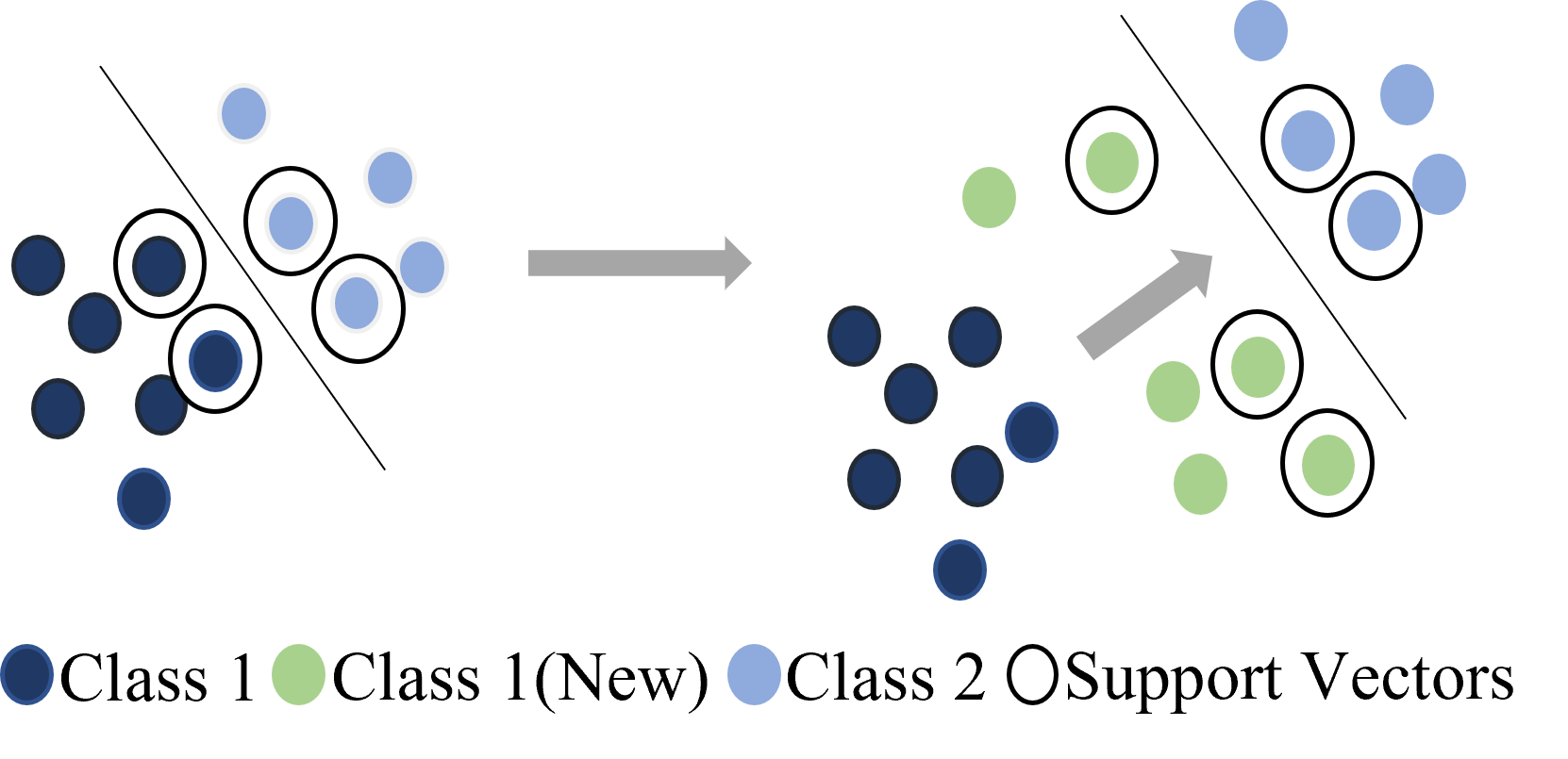}
\caption{Intuition of our subsurface transfer method. We target the support vectors of class~1 (blue circled disks) and generate new samples of class~1 (green) that deliberately shift the boundary (black line) away from the blue class manifold. The other class is presented in light blue.}
\label{fig:sv-augmentation-intuition}
\end{center}
\end{figure}

Building on our previous definitions, we use frequently forgotten regions within the training set to improve robustness and generalization capabilities of the segmentation model. Specifically, we identify frequently forgotten regions within the train set and add example variety through region-wise style transfer. Since we consider forgetting events as an approximate decision boundary proximity metric, our method can be interpreted as an augmentation technique that generates new samples around the class boundaries. Within the manifold, this results in a boundary shift (Figure~\ref{fig:sv-augmentation-intuition}). Based on the popular machine learning paradigm \cite{cortes1995support}, we name our method \emph{Support Vector Augmentation}.

\begin{figure*}[ht]
\begin{center}
\includegraphics[scale=0.6]{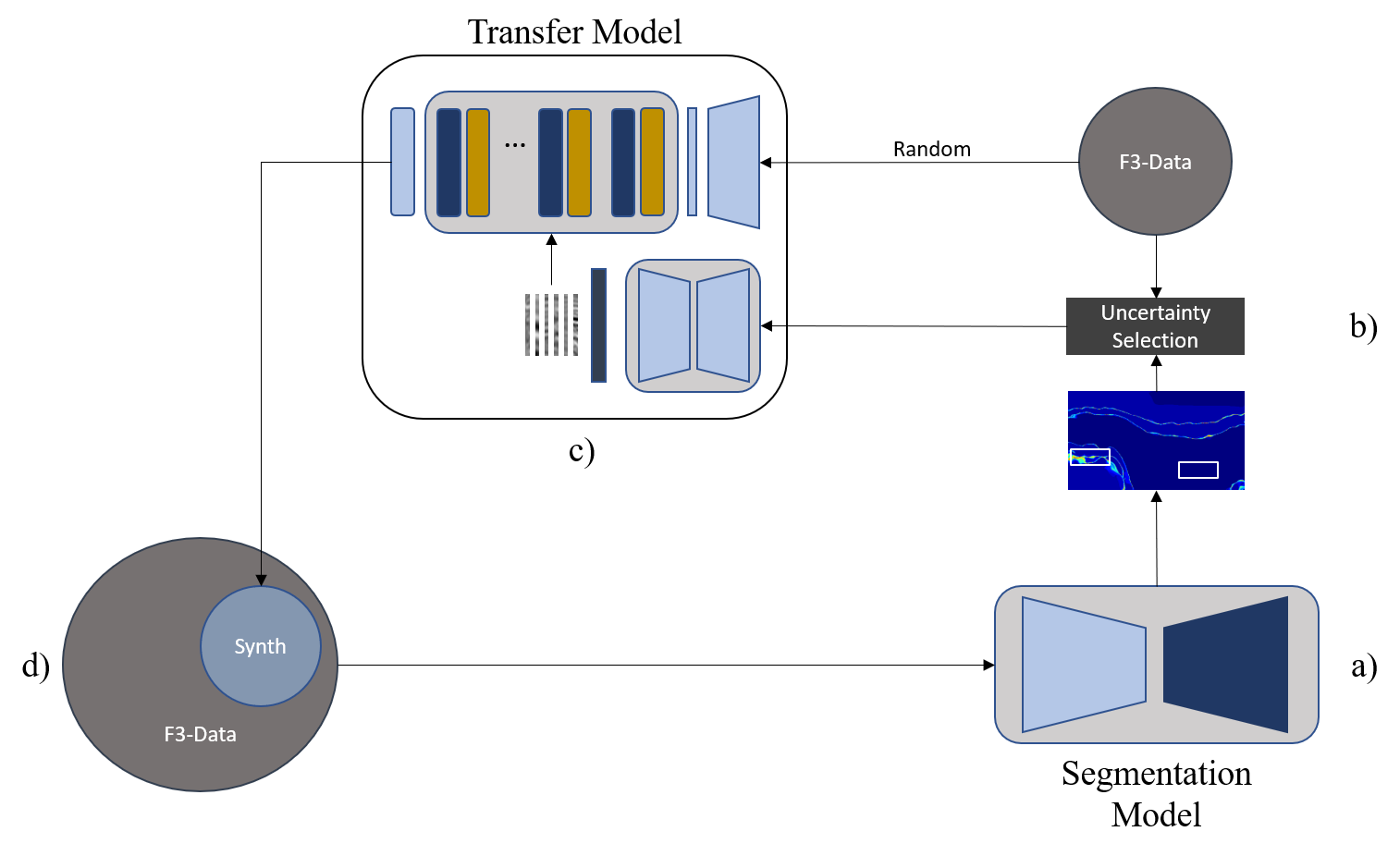}
\caption{The augmentation workflow for seismic interpretation. First, forgetting event maps are produced by the segmentation model using the original volume (a). Second, sections with the most frequently forgotten samples of the target class are selected as the input for our transfer model (b). Third, the transfer model transfers the challenging characteristics of the frequently forgotten regions to other randomly sampled sections throughout the training volume (c). Finally, the generated data is added to the training volume and the segmentation model is retrained from scratch (d).}
\label{fig:subsurface-selection-architecture}
\end{center}
\end{figure*}

Our augmentation workflow consists of a segmentation model, a transfer model and a data selection step (Figure~\ref{fig:subsurface-selection-architecture}). First, our method trains the segmentation model on the training data and produces a forgetting event heat map for every validation image in the training volume (Figure~\ref{fig:subsurface-selection-architecture}a). In principle, heat maps could be produced for the entire training set but that would be computationally inefficient.\\

In the next step of our workflow, we calculate the forgetting event density within each facies class of a heat map (Figure~\ref{fig:subsurface-selection-architecture}b). Specifically, we sum all forgetting events $f_{i\in c_k}$ within class $c_k$ of a heat map and divide by the number of pixels of class $N_{c_k}$ in the image: 

\begin{equation}
U_{c_k} = \frac{\sum_{t=0}^{T}{\sum_{i\in c_k}{f_{i}^{t}}}}{N_{c_k}}.
\end{equation} 
This metric allows us to rank each heat map according to its density with regards to any arbitrary class in the dataset. \\

Finally, we transfer the visual features of a predefined class from the vertical sections with the highest density to randomly sampled training sections (Figure~\ref{fig:subsurface-selection-architecture}c). Our proposed architecture is a slightly altered version of \cite{zhu2020sean}. In short, the model transfers facies characteristics on the batch-normalization activations within the image generator. Our approach enables class specific transfers without affecting the interpretation characteristics (texture, structure etc.) of other classes within the image. In our method, we transfer the underrepresented class facies (e.g. salt domes) due to the learning difficulty of the samples. After generation, the transferred images are added to our training pool and the segmentation model is trained from scratch (Figure~\ref{fig:subsurface-selection-architecture}d). In the remainder of this section, we will discuss our transfer model in more detail as this represents a crucial step in our workflow.\\

\begin{figure*}[!ht]
\includegraphics[scale=0.5]{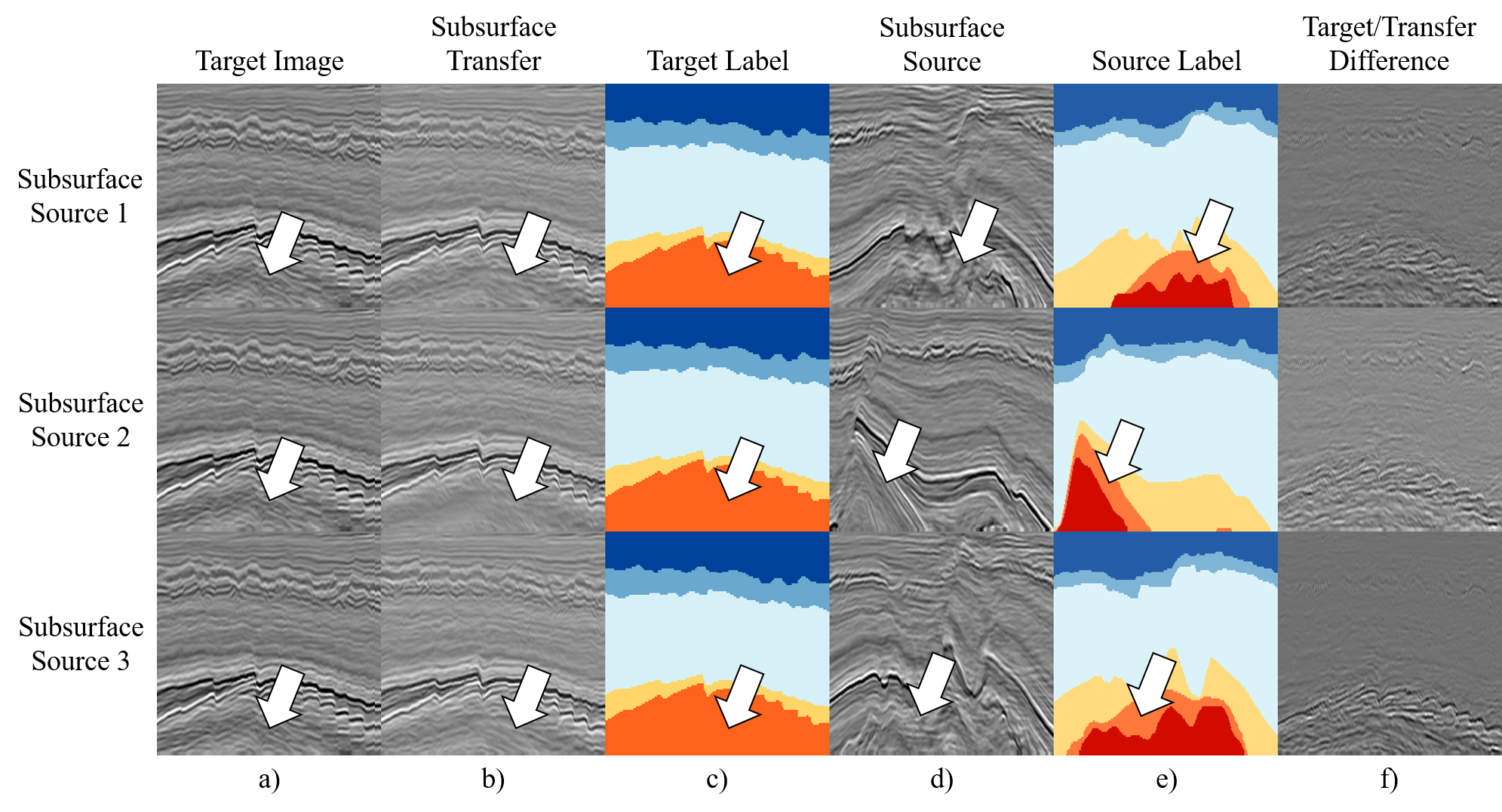}
\caption{Examples of transferred subsurfaces. Columns from left to right: a) Target image for subsurface transfer. b) Output of our algorithm when transferring the orange scruff class. c) Annotation of the target image. d) Subsurface source. e) Subsurface source annotation. f) Absolute differences between adjacent transfer images using different sources (e.g. the difference image of subsurface source~2 represents the difference between the transfer image of the first row and the second row).}
\label{fig:subsurface-transfer}
\end{figure*}

Our transfer model takes three input parameters: A source image, a target image and a class list. The subsurface source and target image are two seismic sections specified by the user. They represent the facies source as well as the section to be altered by our algorithm. The class list contains the facies that our algorithm will transfer to the target image. We show several transfer examples in Figure~\ref{fig:subsurface-transfer}. Here, we transfer the characteristics of the orange scruff class from the source image (Figure~\ref{fig:subsurface-transfer}d) to the target image (Figure~\ref{fig:subsurface-transfer}a) and evaluate the absolute difference between adjacent transfer images using different sources (Figure~\ref{fig:subsurface-transfer}f). For instance, the difference image using source~2 (middle row) represents the absolute difference between the transfer images of the first and second row. As seen in the transfer output, the characteristics of the scruff class are clearly different from the original target image in all example sections. Moreover, we can see that varying the source image results in different scruff facies that are dependent on the scruff regions of their respective source image. For instance, the scruff facie of the transfer image resulting from source~2 is significantly smoother than the other two examples due to the smooth scruff characteristics of subsurface source~2. Finally, the difference plots show what regions are affected by our transfer model when different source images are used. Since our class list only consisted of the scruff class we observe that only this region of the target class is altered by our model. Note, that the difference in the first row (column f) shows the difference between source~1 and source~0, an adjacent source image not shown in our examples.\\

\begin{figure*}[!ht]
\begin{center}
\includegraphics[scale=0.55]{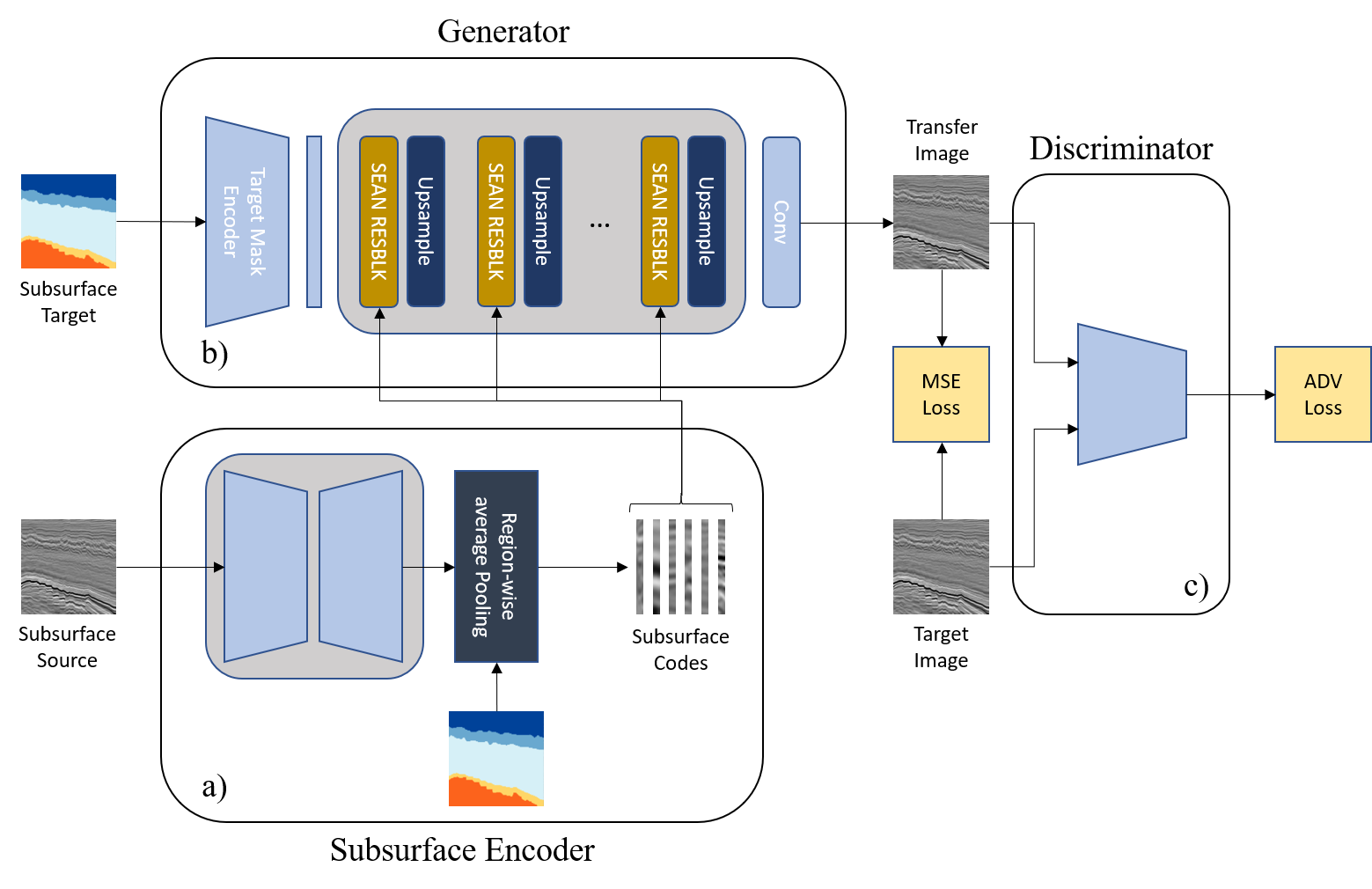}
\caption{The subsurface transfer architecture used in this work during training. Top: Generator with SEAN blocks and upsampling layers (b). Bottom: Subsurface encoder (a) with the bottleneck encoder-decoder architecture (left), region-wise average pooling layer (center) and subsurface codes (right). Left: Discriminator (c) with both loss functions used in training (MSE loss and adversarial loss)}
\label{fig:subsurface-transfer-architecture}
\end{center}
\end{figure*}

To achieve the results in Figure~\ref{fig:subsurface-transfer} we employ a GAN architecture \cite{goodfellow2014generative} consisting of an encoder, a generator as well as a discriminator (Figure~\ref{fig:subsurface-transfer-architecture}). The discriminator (Figure~\ref{fig:subsurface-transfer-architecture}c) predicts whether the presented images originate from the generator or training distribution and is used to derive the adversarial loss. Since this is a standard step in GAN frameworks \cite{goodfellow2014generative} our explanations will focus on the other two architecture elements.\\
We encode subsurface characteristics in two steps (Figure~\ref{fig:subsurface-transfer-architecture}a): First, we encode the source image to remove information irrelevant to subsurface characteristics. Second, we use a class-wise average pooling layer to produce subsurface representations (codes). Each code is a vector and represents the characteristics of one class in the input source. For instance, one code will include the class scruff whereas another code will represent the zechstein class. These codes are used to transfer the facies within the generator. Simply using the target image as the encoder input, will result in an image reconstruction in the generator as the generated codes contain the target image characteristics. However, in our application we want the output to contain the subsurface characteristics of our source image. For this reason, we substitute the subsurface codes with our desired source subsurface codes during inference. This gives us full control over the transfer and allows us to produce examples as seen in Figure~\ref{fig:subsurface-transfer}. In our example, we substitute the subsurface code of the scruff class with the scruff subsurface code from our source image. Note, that we leave all other subsurface codes untouched because we want the other classes to share the same characteristics as the target image.\\

\begin{figure}[!h]
\begin{center}
\includegraphics[scale=0.6]{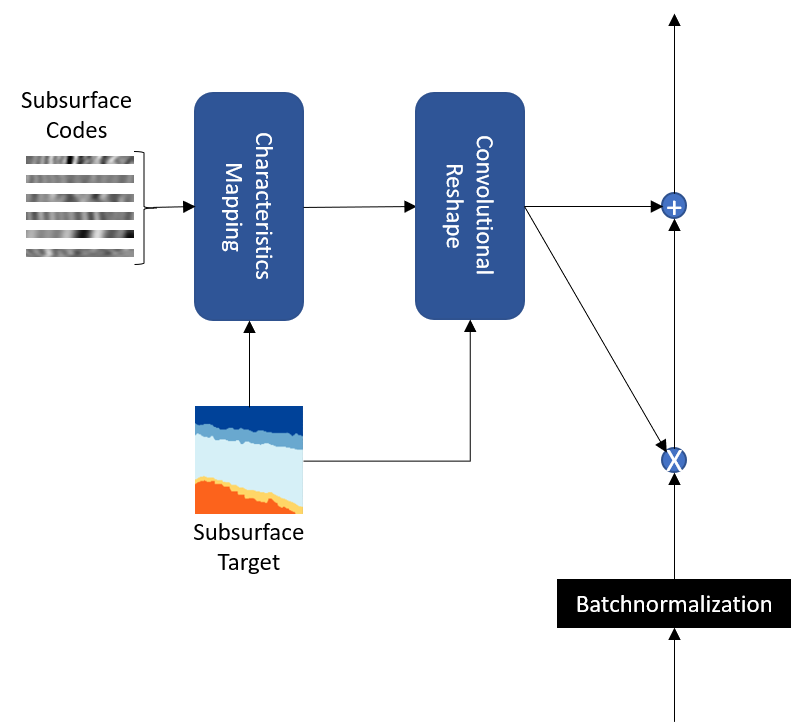}
\caption{SEAN layer workflow. The normalization layer takes the subsurface codes and the target image mask as input parameters (bottom left). The layer modulates the subsurface information as well as the structural information from the annotation onto the normalized output of the previous layer(right).}
\label{fig:sean-block}
\end{center}
\end{figure}

In our generator (Figure~\ref{fig:subsurface-transfer-architecture}b), we use semantic region-adaptive normalization (SEAN) layers to transfer the codes to their respective class regions \cite{zhu2020sean}. In summary, SEAN modulates the subsurface codes as well as the structural information from the annotation onto the normalized output of the previous layer. In Figure~\ref{fig:sean-block} we show a simplified visualization of the modulation process. First, the codes are broadcasted to the respective class regions using the structural information contained in the section annotation (characteristics mapping step). For instance, the scruff subsurface code will be broadcasted to all structural regions containing scruff, the zechstein codes are broadcasted to the zechstein regions and so on. The resulting intermediate image is then used to scale and shift the output of a previous normalization layer.

We train the architecture by learning a simple image reconstruction problem. The subsurface encoder is trained to distinguish per region subsurface codes and the generator is forced to transfer these codes by region adaptive normalization. In inference, the image and annotation source can be different to produce other subsurface codes. In our model, we feed the target image as well as our source image into the encoder sequentially and hand-pick the desired subsurface codes.

\section{Empirical Analysis}
\begin{figure}[!h]
\begin{center}
\includegraphics[scale=0.5]{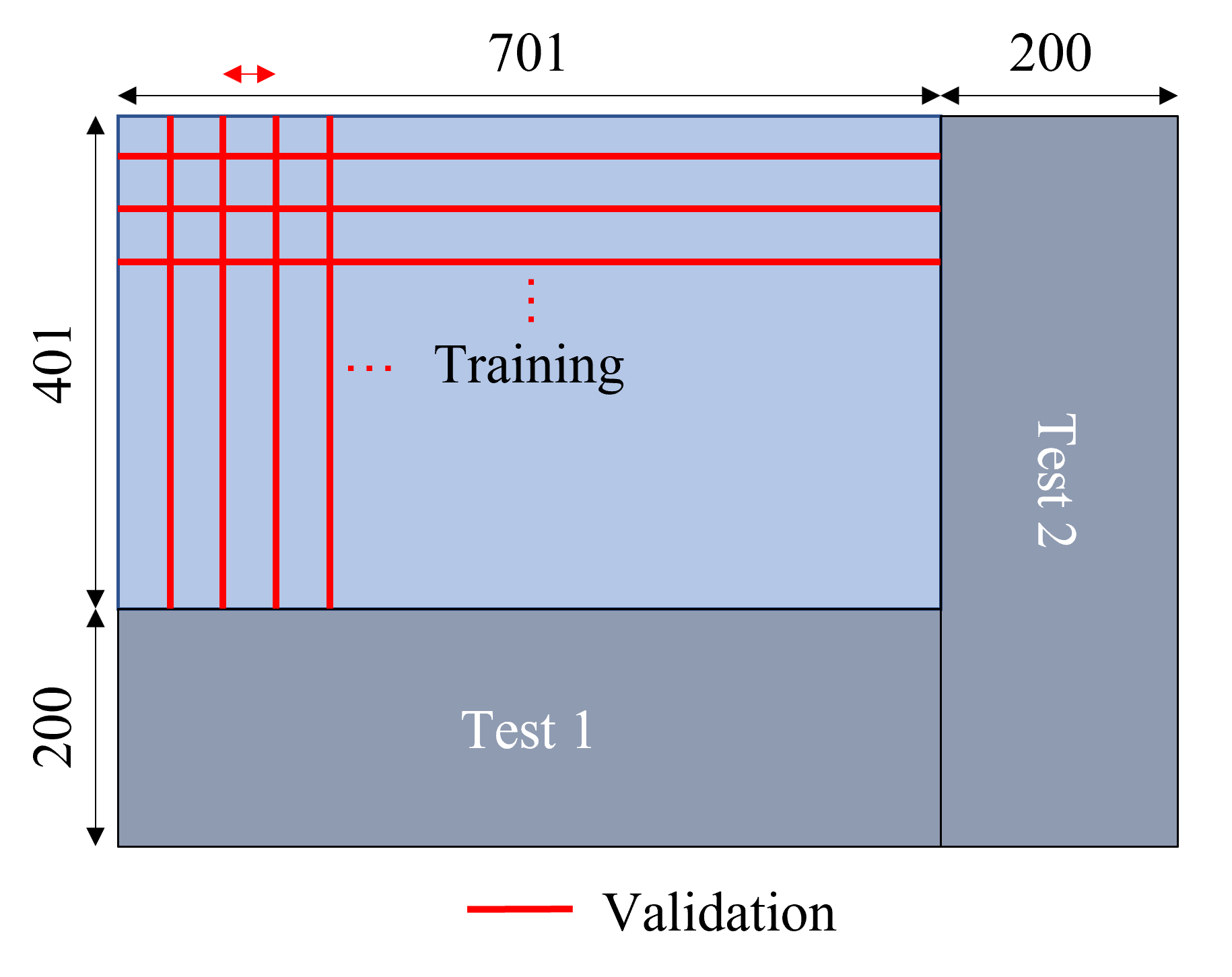}
\caption{Train, test, and validation partition of the F3 dataset. For our validation set, we select every fifth inline and every fifth crossline of the training data.}
\label{fig:partition}
\end{center}
\end{figure}

\begin{figure*}[!ht]
\begin{center}
\includegraphics[scale=0.65]{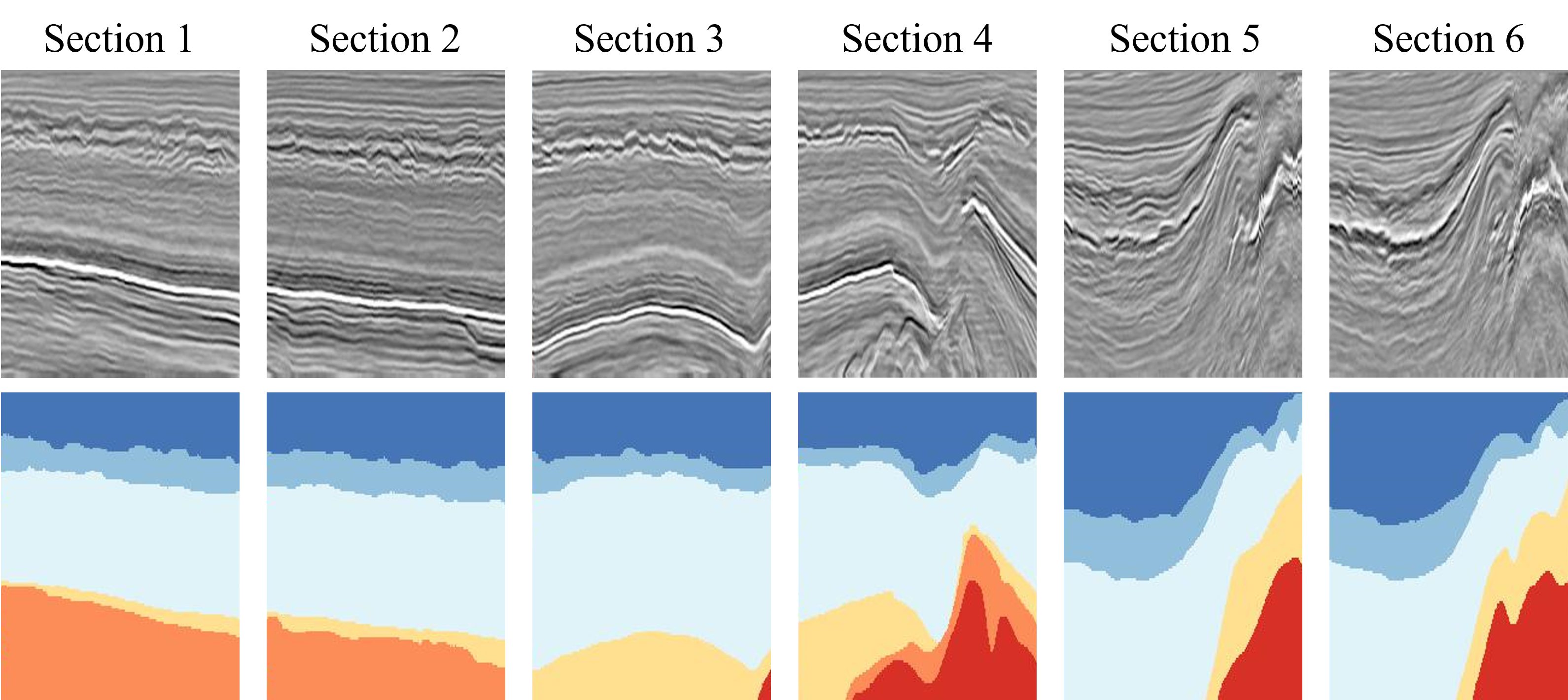}
\caption{Interpreted examples from the test set used for heat map comparison. The sections are chosen based on the high presence of underrepresented classes and complex facies structures.}
\label{fig:interpreted-examples}
\end{center}
\end{figure*}

Our experiments in the entire paper were conducted on the F3 block dataset in the Netherlands \cite{alaudah2019machine}. We partition the volume into train and test set according to the orignal benchmark paper and show the layout in Figure~\ref{fig:partition}. When showing heat maps we restrict the examples to the test sections in Figure~\ref{fig:interpreted-examples}. Here, the sections one through four represent the Test~1 crosslines 234, 310, 556, and 622 respecitvely. Further, sections five and six represent the Test~2 inline sections 575, and 596. Our choice is based on the high presence of underrepresented classes (e.g. the orange scruff class) and complex facies structures. Even though this paper only considers these examples, we note that our observations and conclusions are consistent throughout the entire volume. 

Throughout all of our experiments, we opt for a deeplab-v3 \cite{chen2017deeplab} architecture with different backbone architectures. For optimization, we use the adam variant of stochastic gradient descent with a learning rate of $1e-4$ in combination with a polynomial learning rate decay.
We structure our experiments in two sections: First, we show the analysis benefits of forgetting event heat maps by displaying when pixels are forgotten during training. We distinguish different groups within forgettable pixels and relate them to the model interpretations. Second, we benchmark the generalization and robustness properties of our augmentation method. We analyze the impact of our method along the metrics of segmentation performance and heat map impact. As a comparison we use common augmentation techniques used in literature.

\subsection{Analyzing Forgotten Regions}
\begin{figure*}[ht]
\begin{center}
\includegraphics[scale=0.5]{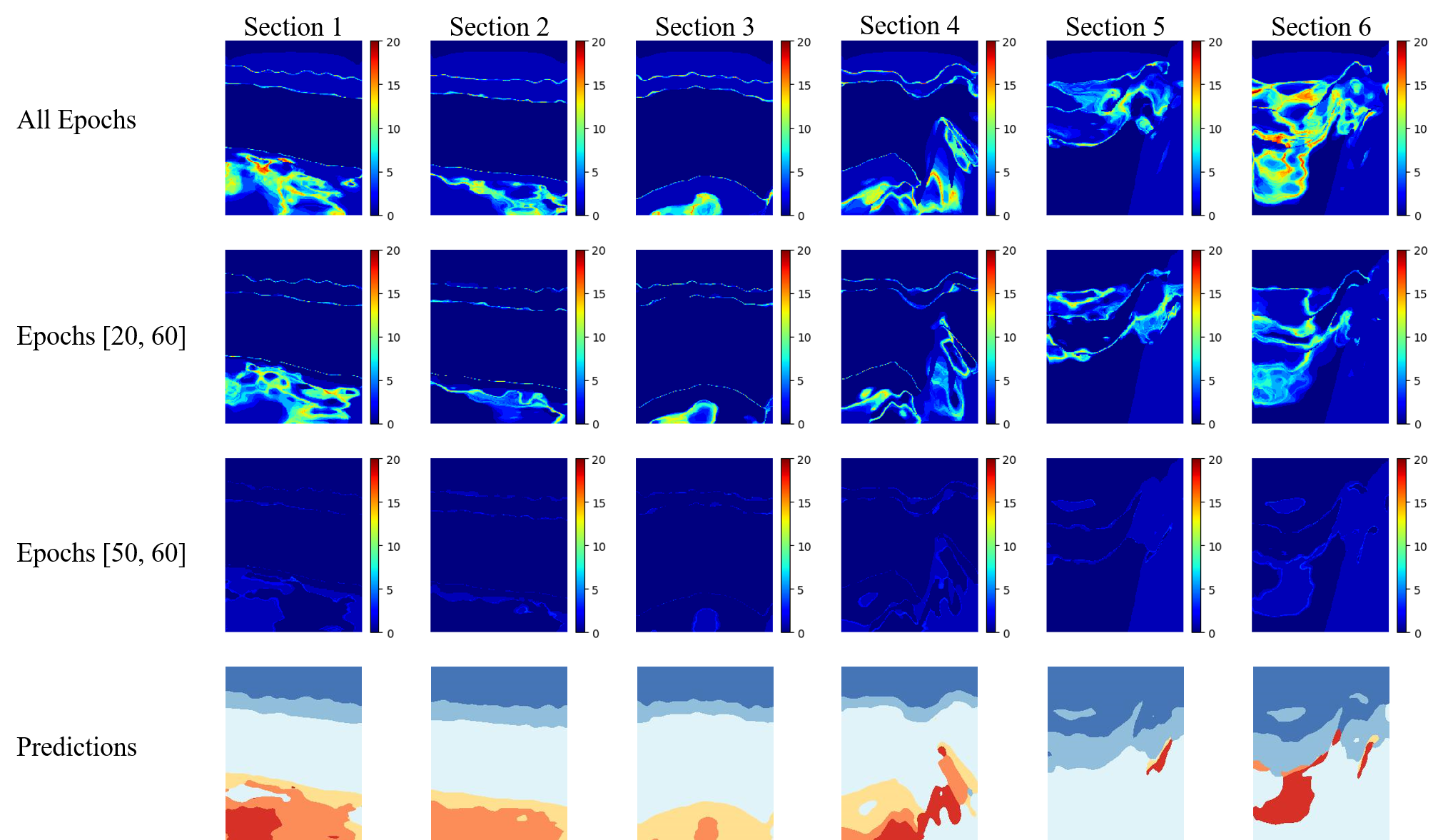}
\caption{Different time frames in which pixels are forgotten. The columns show six different seismic sections from the test set containing difficult textures or structures. Different rows show different observation windows for forgetting events. From top to bottom: 1. Test set heat maps containing all forgetting events throughout training (epoch 0 through epoch 60). 2. Test set heat maps with all forgetting events occurring between epoch 20 and epoch 60. 3. Test set heat maps with all forgetting events occurring between epoch 50 and epoch 60. 4. Predictions of all sections using the fully trained model.}
\label{fig:fevent-recording-analysis}
\end{center}
\end{figure*}

In our experiments, we train the segmentation architecture with a resnet-18 backbone \cite{he2016deep} for 60 epochs on the training volume. We track the forgetting events for the validation and test set and display the heat maps for different observation windows of the test set (Figure~\ref{fig:fevent-recording-analysis}). In each row, we show different time frames in which forgetting events were tracked: In the first row, all forgetting events that occurred during the 60 epochs are displayed. The second and third row show the forgetting events that occurred between the 20th and the 60th epoch as well as the 50th and 60th epoch respectively. In the final third row, we show the predictions of our model. Overall, we can classify forgettable regions in the following groups:\\

The first group consist of pixels forgotten rarely and which disappear from the heat maps after 20 epochs. We call these samples \emph{early-stage forgettable}. These regions are learned at an early stage within the training cycle and the network does not have difficulties mapping them to the representation space. In the context of the interpretation task, these pixels are frequently found in areas that are structurally consistent throughout the volume and do not show a significant variety. An example of these areas can be found in the upper north sea group (dark blue class) and the chaotic middle north sea group (blue class) in images one, two and three. The heat maps within the first row clearly show highlighted regions within the upper classes that disappear after epoch 20 and that are correctly classified by the fully trained model.\\

The second group of pixels is more difficult to characterize by our network and consists of pixels most frequently forgotten - \emph{ambiguous forgettable} samples. These areas are frequently shifted between the correct manifold and other class manifolds during the optimization process but are \emph{not necessarily misclassified}. We interpret these samples to have been within a close proximity to the decision boundary for a specific time frame during the optimization process. Examples of these regions are either the class boundaries or difficult textures within underrepresented classes (e.g. Section~1 frequently forgotten scruff regions in row one and two). Note, these regions are either predicted correctly or incorrectly depending on the network initialization and are not directly visible in the network predicitons.\\

The final group entails the most difficult pixels for our model. This group is consistently classified incorrectly throughout the training procedure and is forgotten rarely at the end of training. Due to this characteristic, we call these samples \emph{late-stage forgettable} samples. In terms of the representation mapping, the network is unable to map these regions into the target manifold throughout training and starts to learn these pixels at a late stage when the model has already learned a large variety of textural and structural features. In our examples, these areas are visible in the third row showing the forgetting events at a late stage. Qualitatively, these areas contain difficult textures or salt dome structures that are not present in the training distribution in that form and hence present the most difficult regions within the test set. This is further confirmed by the false predictions in these regions.

\subsection{Subsurface Transfer}
In this section, we benchmark our support vector augmentation method in combination with common augmentation techniques. Our qualitative heat map results are obtained by training the segmentation model with a resnet-18 backbone for 60 epochs with and without augmentations. For our segmentation performance comparison, we use a resnet-101 backbone and train our model for 80 epochs on five separate random seeds. Furthermore, we train all of our architectures by randomly cropping 255 pixel patches and test with full sections. We choose this setup to ensure a proper comparison to the original baseline of \cite{alaudah2019machine}. For consistency, we choose the validation set by selecting every fifth inline and every fifth crossline of the training volume for all seeds (Figure~\ref{fig:partition}). We note that our numerical results are summarized over every inline and crossline of both Test~1 and Test~2. We query six images with the highest forgetting event density of our target class. Each section is used as a source image to generate 64 transfer images. For generation, we sample randomly to obtain the target image and retrain the segmentation model from scratch. We note that due to the random target image selection our technique is sensitive to the hyperparameter choice (number of sources, targets etc.) and that significant experimentation had to be performed to achieve the results in Table~\ref{table:results-overall}. However, the investigation of different query methods is beyond the scope of this paper and we leave this topic for future research. In this paper, we report the results when transferring the scruff class (orange). For evaluation, we compare our subsurface transfer technique with common augmentation methods (random horizontal flip and random rotations; \cite{orr2003neural}) from computer vision in terms of segmentation performance (in class accuracy and mean intersection over union) and impact on forgetting events heat maps.\\
We show show the numerical results in Table~\ref{table:results-overall}. Overall, every method matches or outperforms the baseline in terms of class accuracy. In particular, our method significantly increases the performance of the target class (scruff) as well as the neighboring underrepresented classes (zechstein and chalk) in the majority of cases. We see that our method affects the accuracy of other classes (upper, middle, and lower north sea group) only mildly and largely remains untouched by the algorithm. For instance, adding our method on top of random rotate increases the scruff class accuracy by 4.5\% while the upper north sea group accuracy is increased by 0.1\% which we consider insignificant. This suggests, that our method is spatially localized and affects the classes in direct proximity of the target class scruff.  

Further, we observe that our method matches the baseline accuracy when combined with the baseline exclusively. Specifically, we observe a maximum of 0.6\% difference to the baseline on the target class as well as its neighboring classes. We reason, that the data variations of our method are not as profound as conventional augmentation techniques and are not as effective when paired with the baseline exclusively. While augmentations such as random rotate result in significant structural variation, our method adds slight subsurface variations to a single class and maintains all other components of the seismic image. Therefore, the augmentation alone does not affect the numerical values strongly. However, when combined with other augmentations its effect becomes amplified and more pronounced. For instance, we see a clear improvement when using our method in combination with random rotate.

Finally, we note that adding an augmentation can result in minor accuracy reductions for selective classes. For instance, adding random rotate to the baseline results in a 1.3\% reduction in terms of accuracy on the upper north sea class. While augmentations frequently result in an overall accuracy improvement, several augmentations can have a negative effect on specific class groups or even entire section performances. In the example of random rotate, the upper north sea class does not share an upper boundary with another class and is therefore "cut off" when rotated. However, we note that adding our method does not result in such a behavior and reductions can be considered irrelevant. This affirms that our method introduces realistic data variations for seismic interpretation for every class and therefore matches or improves the baseline performance.

\begin{figure*}[!h]
\begin{center}
\includegraphics[scale=0.52]{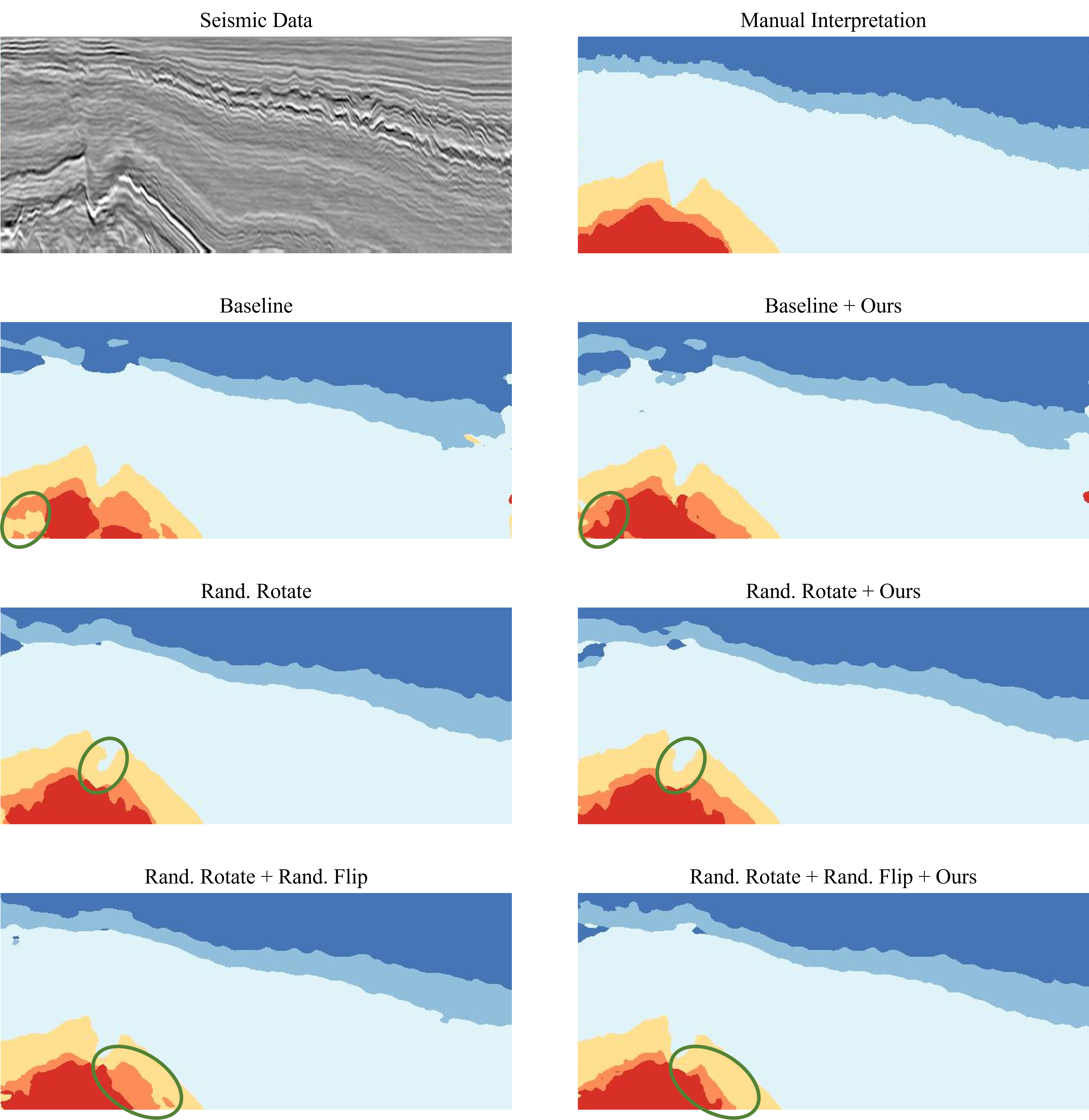}
\caption{Model predictions for crossline 60 in Test~2 when different augmentation methods are used. Top row: Seismic data and the corresponding ground truth predictions.Rows two, three, and four: No augmentations (Baseline), random rotate, and random rotate as well as random horizontal flip. The first column contains the augmentation, and the second column contains the same augmentation with our method. Areas of improvmenta are highlighted in green circles.}
\label{fig:predictions}
\end{center}
\end{figure*}

In addition to Table~\ref{table:results-overall}, we further show the predictions of crossline 60 in Test~2 for different augmentation constellations (Figure\ref{fig:predictions}). We highlight areas of improvement in green. Overall, the predictions further support our numerical analysis of Table~\ref{table:results-overall}. In particular, we find that adding our method in any constellation typically results in more fine-grained predictions that are less smooth. For instance the highlighted area when using random rotate and random flip contains significantly smoother scruff predictions than the model trained with our method. We reason, that our method introduces style variations into the data that provide more boundary robustness. For this reason, the predictions are more fine-grained.\\

\begin{figure*}[!h]
\begin{center}
\includegraphics[scale=0.65]{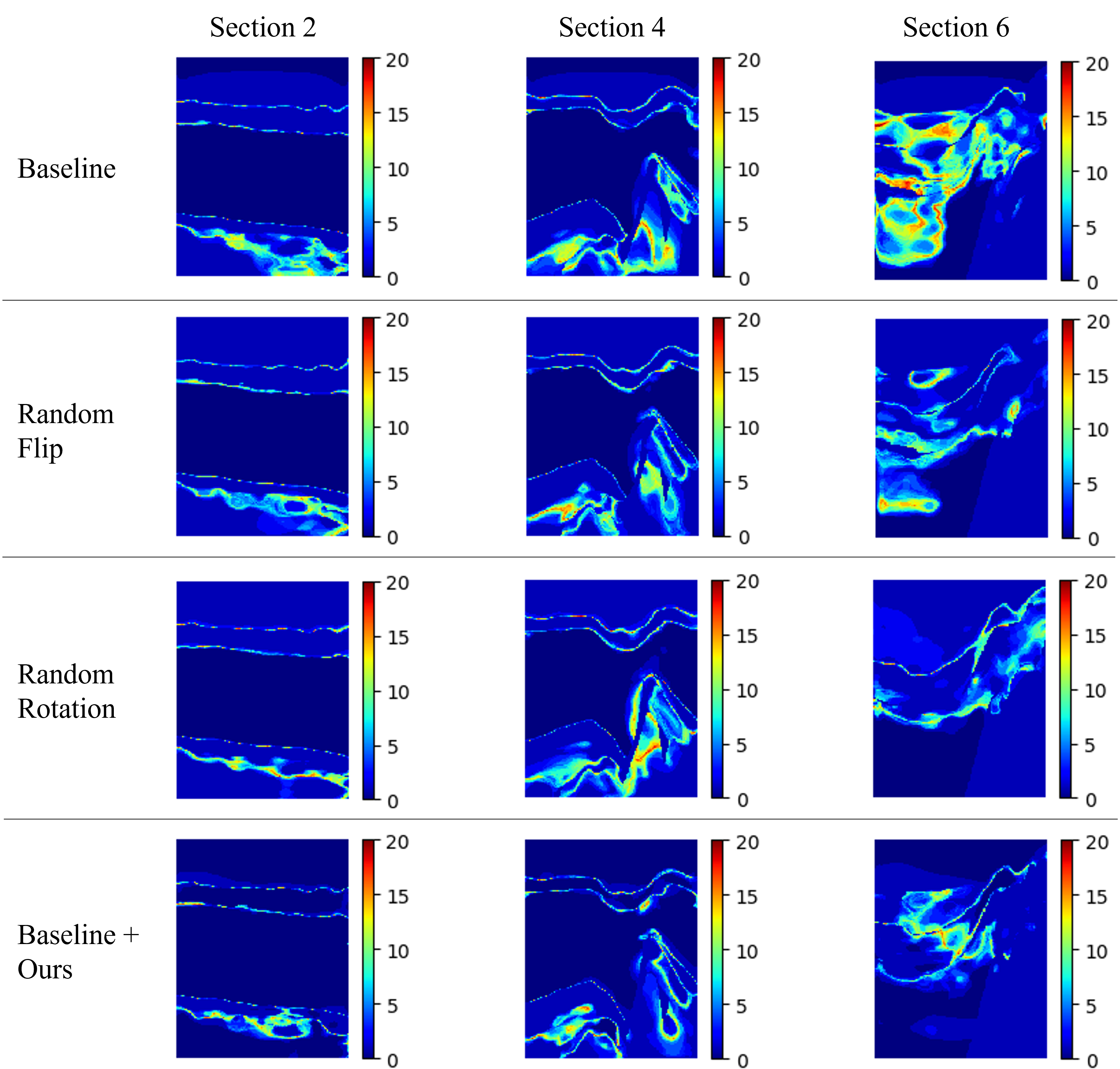}
\caption{Heat maps for the example sections two, four, and six when using different augmentation methods. Each group represents a different augmentation method: The first row contains the baseline model with no augmentations. The second row contains random horizontal flip where we flip each section randomly during training. The third row contains randomly rotated image augmentations. The final row represents our subsurface augmentation method. In summary, our method reduces the amount of forgetting events significantly more than traditional methods.}
\label{fig:sv-augmentation}
\end{center}
\end{figure*}

We further show the forgetting event heat maps of the different augmentations in Figure~\ref{fig:sv-augmentation}. Qualitatively, our method reduces the amount of forgetting events significantly more than traditional augmentation methods indicating a clear representation shift. Specifically, we find that several regions with a high forgetting event density are transferred to a low forgetting event density or disappear entirely (bottom scruff class in Section~2, entire left part of Section~6, or center of Section~4). These regions are shifted away from the decision boundary and the classification accuracy is not significantly affected by model updates. In contrast, forgettable regions do not disappear with standard augmentation techniques. Instead, only the severity of the forgetting event regions is reduced or the texture of the regions is blurred. For instance, random rotation results in blurred edges around the forgettable regions.\\
Finally, we identify regions that transition from lower forgetting event densities to higher densities (e.g. Section~6 bottom left) when using our augmentation method. Because these regions transition to more difficult regions, and serve as an example of negative representation shifts. However, we also note empirically that these occasions are rare and that the reduction of forgettable regions is significantly more common than an increase.

\begin{table*}[!ht]
	\centering
	\begin{tabular}{ |p{2.3cm}||p{1.5cm}|p{1.5cm}|p{1.5cm}|p{1.5cm}|p{1.5cm}|p{1.5cm}|p{1.5cm}|}
		\hline
		\multicolumn{8}{|c|}{Class Accuracy \& Overall MIoU} \\
		\hline
		Model & MIoU & Upper N. S. & Middle N. S. & Lower N. S. & Chalk & Scruff & Zechstein\\
		\hline
		Baseline & 0.689 & 0.986 & 0.875 & 0.965 & 0.771 & $\mathbf{0.591}$ & $\mathbf{0.622}$\\
		\hline
		Baseline + Ours & 0.687 & 0.983 & 0.886 & 0.962 & $\mathbf{0.772}$ & 0.585 & 0.619\\
		\hline
		Rand. Rotate & 0.709 & 0.973 & 0.923 & 0.972 & 0.792 & 0.600 & 0.556\\
		\hline
		Rand. Rotate + Ours & 0.724 & 0.974 & 0.930 & 0.973 & $\mathbf{0.811}$ & $\mathbf{0.646}$ & $\mathbf{0.593}$\\
		\hline
		Rand. Rotate + Rand. Flip & 0.728 & 0.973 & 0.927 & 0.974 & 0.804 & 0.650 & 0.588\\
		\hline
		Rand. Rotate + Rand. Flip + Ours & 0.732 & 0.971 & 0.927 & 0.975 & $\mathbf{0.822}$ & $\mathbf{0.664}$ & $\mathbf{0.641}$\\
		\hline
	\end{tabular}
	\caption{Averaged class accuracy.}
	\label{table:results-overall}
\end{table*}

\section{Conclusion}
In this paper, we presented a novel framework that enhances explainability in deep seismic models. We track the frequency in which pixels are forgotten during training and analyze the relationship to the sample position within the feature space. We highlight forgotten pixels spatially in heat maps and interpret their semantic geologic meaning. Further, we consider different time frames in which samples are forgotten and are able to tie specific prediction properties to model behaviour. Finally, we exploit our framework to engineer an augmentation method that explicitly targets forgotten regions and increases the variety of difficult pixels through subsurface transfer. Our empirical evaluations clearly show the shift in the learned feature space when compared to traditional augmentation methods. In future, we hope that this work will provide a powerful concept for interpreters to verify the model functionality and explain its behaviour. Furthermore, we have shown that the well crafted methods can target prone regions and allow explicit control over the decision boundary. Future work could include an application exploration and forgetting events could be applied to multiple seismic applications such as rock lithology predictions or salt body delineation.

\bibliographystyle{IEEEtran}
\bibliography{mybib}

\end{document}